\documentclass[11pt,a4paper]{article}
\usepackage[hyperref]{acl2021}
\usepackage{times}
\usepackage{latexsym}

\usepackage{microtype}

\usepackage{xcolor}
\usepackage{color, colortbl}
\usepackage{comment}
\usepackage{xspace}
\usepackage{multirow}
\usepackage{array}
\usepackage{float}
\usepackage{booktabs}
\usepackage{tikz}
\usepackage{caption}
\usepackage{enumitem}
\usepackage{tabu}
\usepackage{rotating}
\usepackage{amssymb}
\usepackage{amsmath}
\usepackage{tikz}
\usepackage{microtype}
\usepackage{capt-of,etoolbox}

\aclfinalcopy %

\title{\textsf{\textsc{Dodrio}}: Exploring Transformer Models with Interactive Visualization}

\author{
    Zijie J. Wang
    \quad Robert Turko
    \quad Duen Horng (Polo) Chau \vspace{2pt}\\
    College of Computing, Georgia Tech \vspace{2pt} \\
    \texttt{\{jayw, rturko3, polo\}@gatech.edu}
}

\date{}

\makeatletter
\let\@oldmaketitle\@maketitle%
\renewcommand{\@maketitle}{\@oldmaketitle%
  \vspace{-18mm}
  \includegraphics[width=\linewidth]{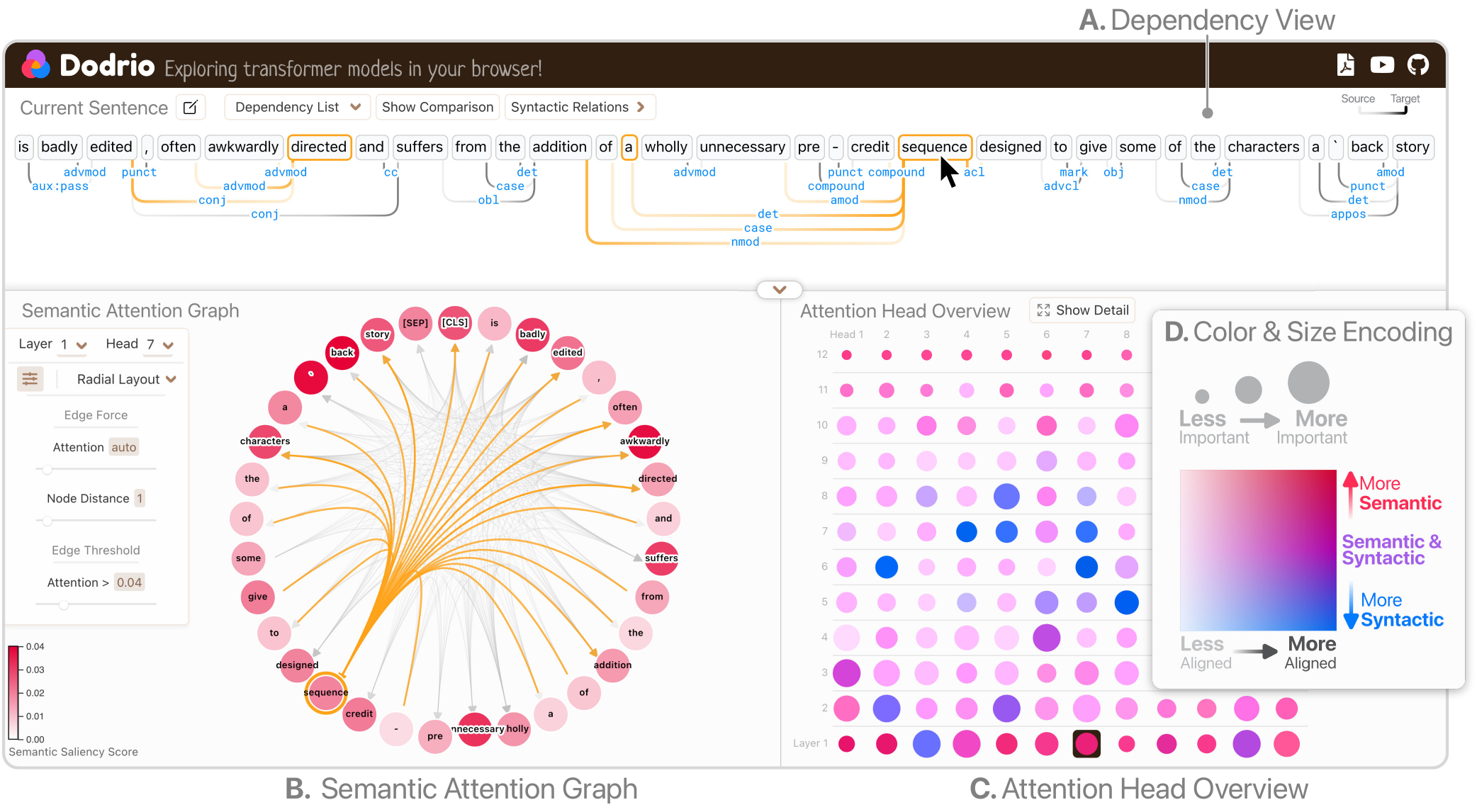}
  \vspace{-7mm}
  \captionof{figure}{
    The \tool{} user interface showing user exploration of connections between attention weights from a fine-tuned BERT model and syntactic dependencies as well as semantic saliency scores on the SST2 dataset.
    \textbf{(A)} \textbf{\dependencyView{}} enables users to hover over a word from the input sentence to highlight its associated dependency directed links as \textcolor{orange}{\textbf{orange}} arcs (\textcolor{lightorange}{\textbf{lighter}} is \textit{source}; \textcolor{darkorange}{\textbf{darker}} is \textit{target}).
    \textbf{(B)} \textbf{\attentionMapView{}} highlights the word's related tokens and their attentions; 
    nodes are tokens (darker means more salient); a directed edge encodes attention weight between two tokens.
    \textbf{(C)} \textbf{\overview{}} shows all attention heads in a multi-layer and multi-head model as a grid of circles, each head is
    \textbf{(D)} 
    \textbf{colored} based on its linguistic knowledge in the model (more \textcolor{red}{\textbf{red}}$\rightarrow$more semantic-aligned, more \textcolor{blue}{\textbf{blue}}$\rightarrow$more syntactic-aligned; darker$\rightarrow$more aligned), 
    and \textbf{sized} based on its importance score in the model (larger$\rightarrow$more important).
  }
  \label{fig:overview}
  \vspace{5mm}
 }
\makeatother

\definecolor{orange}{HTML}{EB9130}
\definecolor{lightorange}{HTML}{F4C28D}
\definecolor{darkorange}{HTML}{C46F13}
\definecolor{red}{HTML}{FF0E58}
\definecolor{agreen}{RGB}{74, 198, 148}
\definecolor{purple}{HTML}{A561DF}
\definecolor{blue}{HTML}{007FFA}
\definecolor{darkpurple}{RGB}{170, 70, 210}
\definecolor{aqua}{RGB}{87, 180, 181}
\definecolor{lightblue}{RGB}{72, 123, 232}
\definecolor{hotpink}{RGB}{255, 83, 115}
\definecolor{teal}{RGB}{90, 200, 250}
\definecolor{linkColor}{RGB}{6,125,233}

\definecolor{white}{HTML}{FFFFFF}
\definecolor{black}{HTML}{000000}

\newcommand{\tool}{\textsc{Dodrio}}
\newcommand{\embeddingView}{Embedding View}
\newcommand{\tableView}{Table View}
\newcommand{\overview}{Attention Head Overview}
\newcommand{\attentionMapView}{Semantic Attention Graph}
\newcommand{\dependencyView}{Dependency View}

\newcommand{\initialDependencyView}{Dependency View}
\newcommand{\expandedComparisonView}{Comparison View}
\newcommand{\instanceSelectionView}{Instance Selection View}

\begin{document}

\maketitle

\begin{abstract}

Why do large pre-trained transformer-based models perform so well across a wide variety of NLP tasks?
Recent research suggests the key may lie in multi-headed attention mechanism's ability to learn and represent linguistic information.
Understanding how these models represent both syntactic and semantic knowledge is vital to investigate why they succeed and fail, what they have learned, and how they can improve.
We present \tool{}, an open-source interactive visualization tool to help NLP researchers and practitioners analyze attention mechanisms in transformer-based models with linguistic knowledge.
\tool{} tightly integrates an overview that summarizes the roles of different attention heads, and detailed views that help users compare attention weights with the syntactic structure and semantic information in the input text.
To facilitate the visual comparison of attention weights and linguistic knowledge, \tool{} applies different graph visualization techniques to represent attention weights scalable to longer input text.
Case studies highlight how \tool{} provides insights into understanding the attention mechanism in transformer-based models.
\tool{} is available at \url{https://poloclub.github.io/dodrio/}.

\end{abstract} %
\section{Introduction}

The rise of transformer-based models have brought dramatic performance improvements across many NLP tasks~\cite{wangGlue2019}.
In particular, BERT~\cite{delvinBert2019} has demonstrated that transformer-based models pre-trained on large-scale corpora can be effectively fine-tuned for a wide variety of downstream tasks, such as sentiment analysis, question answering, and text summarization.
However, how these language models generalize text representations learned from an unsupervised training process to downstream sentence understanding tasks remains unclear.
There is a growing research body in interpreting transformer-based models, as understanding what these models have learned and why they succeed and fail is vital for NLP researchers to develop better models, and critical for decision makers to trust these models.

The current approach on interpreting transformer-based models focuses on probing and attention weight analysis~\cite{hewitt-liang-2019-designing}.
There is an active discussion on whether attention weights are explanations~\cite{jainAttentionNotExplanation2019b}, but more recent work has shown that they do provide insights on what the models have learned~\cite{atanasovaDiagnosticStudyExplainability2020b}.
In particular, research has shown that transformer-based models have learned to represent semantic knowledge and lexical structure in text~\cite{rogersPrimerBERTologyWhat2020a}.
Furthermore, interaction visualization systems have shown great potential in explaining complex deep learning models~\cite{hohman2018visual, wangCNNExplainerLearning2020}.
Some visualization tools have been developed for transformer-based models~\cite{vigAttentionVis2019, hooverExBERTVisualAnalysis2020, deroseAttentionFlowsAnalyzing2021}.
However, these systems usually focus on visualizing and analyzing attention weights, instead of visually connecting them to linguistic knowledge that is crucial to investigate why transformer-based models work so well across different tasks~\cite{rogersPrimerBERTologyWhat2020a}.

To address this research challenge, we present \textbf{\textsf{\tool{}}} (\autoref{fig:overview}), an interactive visualization tool to help NLP researchers and practitioners analyze and compare attention mechanisms with linguistic knowledge.
For a demo video of \tool{}, visit \url{https://youtu.be/qB-T9j7UTgE}.
In this work, our primary contributions are:
\begin{enumerate}[topsep=2mm, itemsep=0.6mm, parsep=0mm, leftmargin=4mm]
    \item \textsf{\textbf{\tool{}},} \textbf{a novel interactive visualization system} that helps users better understand the attention mechanisms in transformer-based models by linking attention weights to semantic and syntactic knowledge.
    \item \textbf{Novel interactive visualization design} of \tool{}, which integrates overview + detail, linking + brushing, and graph visualizations that simultaneously summarizes a complex multi-layer and multi-head transformer model, and provides linguistic context for users to interpret attention weights at different levels of abstraction.
    \item \textbf{An open-source}\footnote{\url{https://github.com/poloclub/dodrio}} and \textbf{web-based implementation} 
    that broadens the public’s access to modern deep learning techniques.
    We also provide thorough documentations to encourage users to extend \tool{} to their own models and datasets.
\end{enumerate}

\section{Background}

Attention heads are comprised of weights incurred from words when calculating the next representation of the current word~\cite{clarkWhatDoesBERT2019a}, which are known as attention weights.
Easily interpretable, using attention to understand model predictions across domains is a very popular research area~\cite{xuShowAttend2020, rocktaschelReasoningAboutEntailment2016}.
In NLP, there has been a growing body of research on attention used as a tool for interpretability across many language tasks~\cite{wiegreffeAttentionNotNot2019, vashishthAttentionInterpretabilityNLP2019, kobayashiAttentionNotOnlyWeight2020}.

Existing visualization systems and techniques do not visually connect attention mechanisms to linguistic knowledge~\cite{tenneyLanguageInterpretabilityTool2020, deroseAttentionFlowsAnalyzing2021}, we propose novel visualization approaches that foster exploration across semantically and syntactically significant attention heads in complex model architectures.
For example, for every attention head in the 144 heads of BERT, the entry $A_{i,j}$ in the attention map $A$, represents the attention weight from token $i$ to token $j$. With 144 $\times$ number of tokens $\times$ number of tokens attention weights in BERT for each input instance, it is challenging to systematically analyze these attention weights without abstraction and linguistic context.
\tool{} aims to address this challenge by applying novel interactive visualization techniques.
\section{Interface}

\subsection{\overview{}}
As a user explores the attention weights, the \overview{} (\autoref{fig:overview}C) serves as a guide to effectively navigate the remaining views of the interface.
With visual linking and brushing~\cite{mcdonaldOrion1988}, we unify attention head selection with the state of the remainder of the interface.
This view of a grid of attention heads guides the user to inspect semantically and syntactically important heads.
Attention heads are encoded as circles where color encodes the head's linguistic alignment (more red$\rightarrow$more semantic-aligned, more blue$\rightarrow$more syntactic-aligned; darker$\rightarrow$more aligned), and sized represents its importance score in the model (larger$\rightarrow$more important) (\autoref{fig:overview}B).

We calculate the \textbf{semantic score} $m$ by computing the cosine similarity between the sum of attentions received for each token at a given head, and the sentiment score of each token.
If the sentiment score is not available in a dataset, we use the saliency score for each token instead.
The saliency score of a token measures how important that token contributes to the final model prediction~\cite{BARREDOARRIETA202082}, and it is shown to correlate with word semantics~\cite{atanasovaDiagnosticStudyExplainability2020b}.

Following \citet{clarkWhatDoesBERT2019a}'s framework, we use the source token's most-attended token as its predicted dependency target.
For each existing dependency relationship, we compute each head's average accuracy across all instances.
Finally, we calculate the head's \textbf{syntactic score} $n$ by taking the maximum of its average accuracy across all existing dependency relationships (ground truth or generated by a parser).

There are multiple metrics to measure the importance of a given attention head.
By default, we calculate the \textbf{importance score} $c$ of an attention head by the average of its maximum attention for all instances in the dataset~\cite{voitaSpecializedHeads2019}.
\tool{} also supports using the sum of absolute gradients of attention weights in an attention head as its importance score $c$~\cite{clarkWhatDoesBERT2019a}.

\begin{figure}[tb]
    \includegraphics[width=\linewidth]{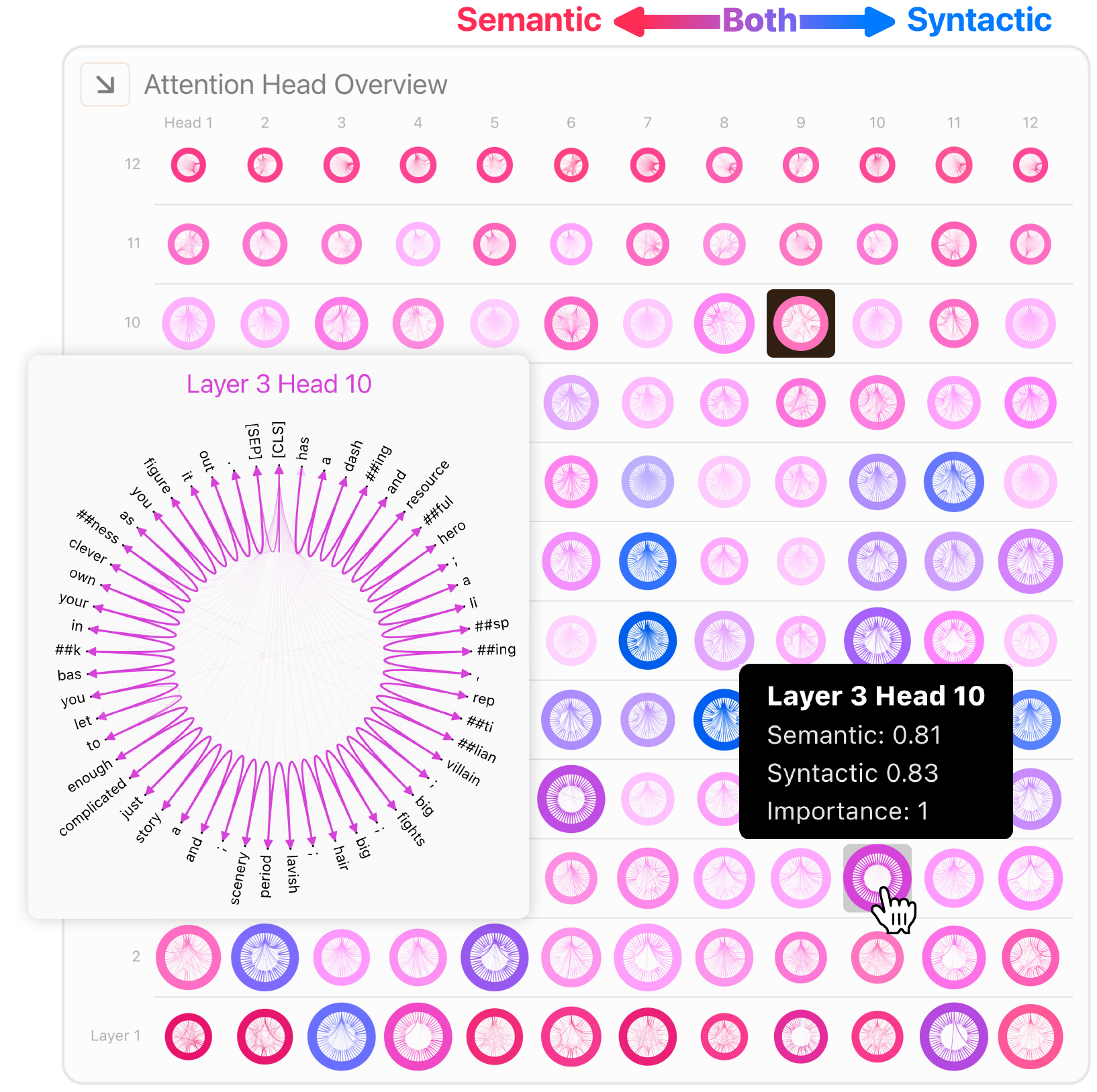}
    \caption{
     The expanded \overview{} provides a preview of all attention heads for the input sentence.
     Attention heads are represented as a grid of rings (right) where their attention weights are shown in the middle.
     Each ring's color and size encode the attention head's linguistic knowledge alignment and importance score (\textcolor{red}{{\textbf{red}}}$\rightarrow$semantic; \textcolor{purple}{{\textbf{purple}}}$\rightarrow$semantic and syntactic; \textcolor{blue}{{\textbf{blue}}}$\rightarrow$syntactic; larger$\rightarrow$more important).
     Users can click an attention head to inspect its attention weights in detail in a radial layout window (left).
     \looseness=-1
    }
    \label{fig:atlas}
\end{figure}

After computing these three scores, we create a linear color scale and a linear size scale to encode them in the \overview{} (\autoref{fig:overview}C, D).
We use the Hue-chroma-luminance (HCL) color space to represent colors in \tool{}.
The HCL color space is designed to better align with human perception of colors, so that interpolations in this space is smoother and more consistent~\cite{zeileis2009escaping}.
We use the hue value (H) in the HCL color space to encode $m - n$ with range [-1, 0, 1] as [\textcolor{blue}{{\textbf{blue}}}, \textcolor{purple}{{\textbf{purple}}}, \textcolor{red}{{\textbf{red}}}]; the luminance value (L) to encode $\max\left(m, n\right)$ (range [0, 1]); and the size of circles to encode $c$ (range [0, 1]).
With our color and size encoding, the \overview{} (\autoref{fig:overview}C and \autoref{fig:atlas}) provides an accurate and efficient summarization of attention heads.\looseness=-1

In the \overview{}, users can also click a button to show the expanded \overview{} (\autoref{fig:atlas}) that additionally provides a preview of the attention pattern in each attention head through the \textit{Radial Layout} visualization.
Hovering over one attention head displays its linguistic and importance information.

\subsection{Syntactic Dependencies}
Word relations in a sentence are important features to understand the lexical makeup of a sentence, which can help users further deduce model decisions in the context of sentence structure.
In \tool{}, a user can explore an attention head with input sentence's dependency relationships.

\textbf{\initialDependencyView{}} (\autoref{fig:overview}A). 
We visualize true dependency relations, if available, or relations tagged by the CoreNLP pipeline~\cite{manningCoreNLP2014} linked with the \attentionMapView{} for users to investigate syntax-sensitive behavior at different attention heads.
The user can further explore the dependency representation in a hierarchical structure by filtering dependency relations.

\begin{figure*}[h]
    \includegraphics[width=\linewidth]{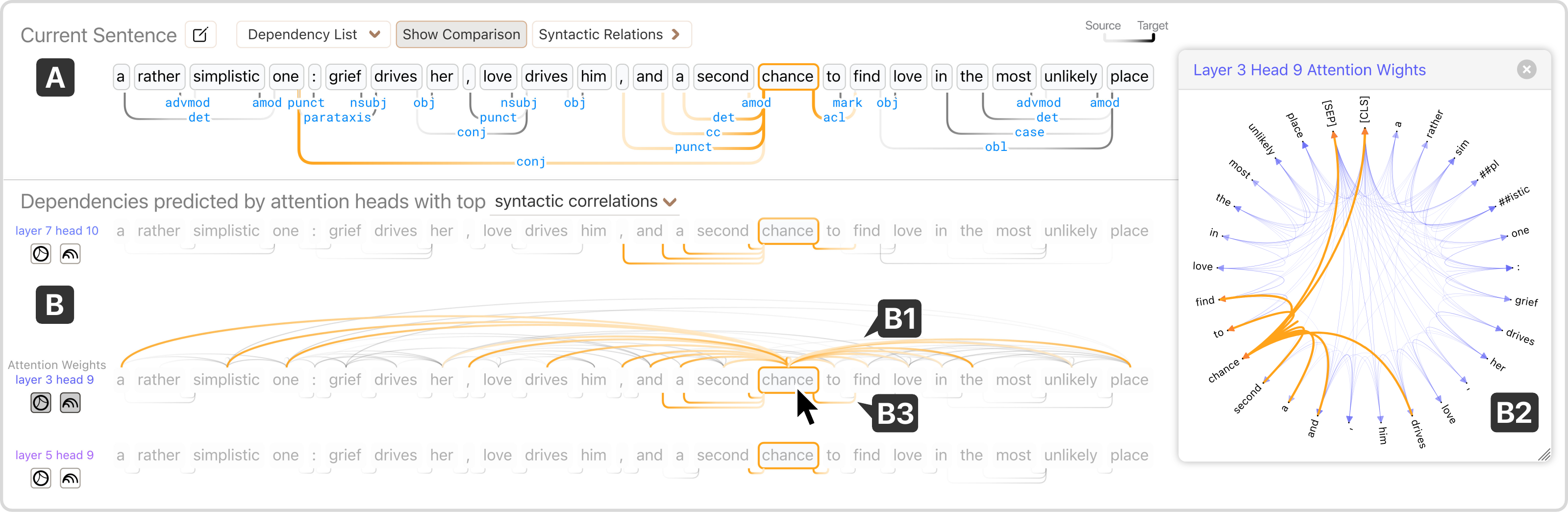}
    \caption{
    The \expandedComparisonView{} allows users to compare multiple attention heads and explore the connection between attention weights and the syntactic structure of the input sentence.
    (\textbf{A}) The top \textit{rectangular arc diagram} visualizes dependencies generated by a parser (lighter is source; darker is target).
    (\textbf{B}) Each attention head is represented as a row of tokens where (\textbf{B1}) the top \textit{curved arc diagram} and (\textbf{B2}) the \textit{radial layout window} display the selected head's attention weights on demand.
    (\textbf{B3}) The \textit{rectangular arc diagram} below the tokens shows the dependencies predicted using attentions.
    Hovering over one token highlights all associated attentions and dependency links.\looseness=-1
    }
    \label{fig:comparison}
\end{figure*}

\textbf{\expandedComparisonView{}} (\autoref{fig:comparison}).
Understanding raw attention weights are best interpreted relative to the attention weights at other attention heads in the model.
The \expandedComparisonView{} enables users to examine the dependencies predicted by attention heads (\autoref{fig:comparison}-B3).
A user can select additional attention representations under each attention head label within this view to supplement their analysis of attention with respect to the grammatical structure of the sentences.
By viewing the attention edges drawn above the tokens, which encode attention weight magnitude with opacity \textit{in the Arc Layout}  (\autoref{fig:comparison}-B1), a user can maintain word-order context in the sentence, while the attention representation utilizing a \textit{Radial Layout} (\autoref{fig:comparison}-B2) of attention edges allows for a clearer interpretation the attention distribution.
The edge linking with interaction between this view and the \initialDependencyView{} further reinforces the syntax-sensitive behavior present in attention heads
\begin{figure}[t]
    \includegraphics[width=\linewidth]{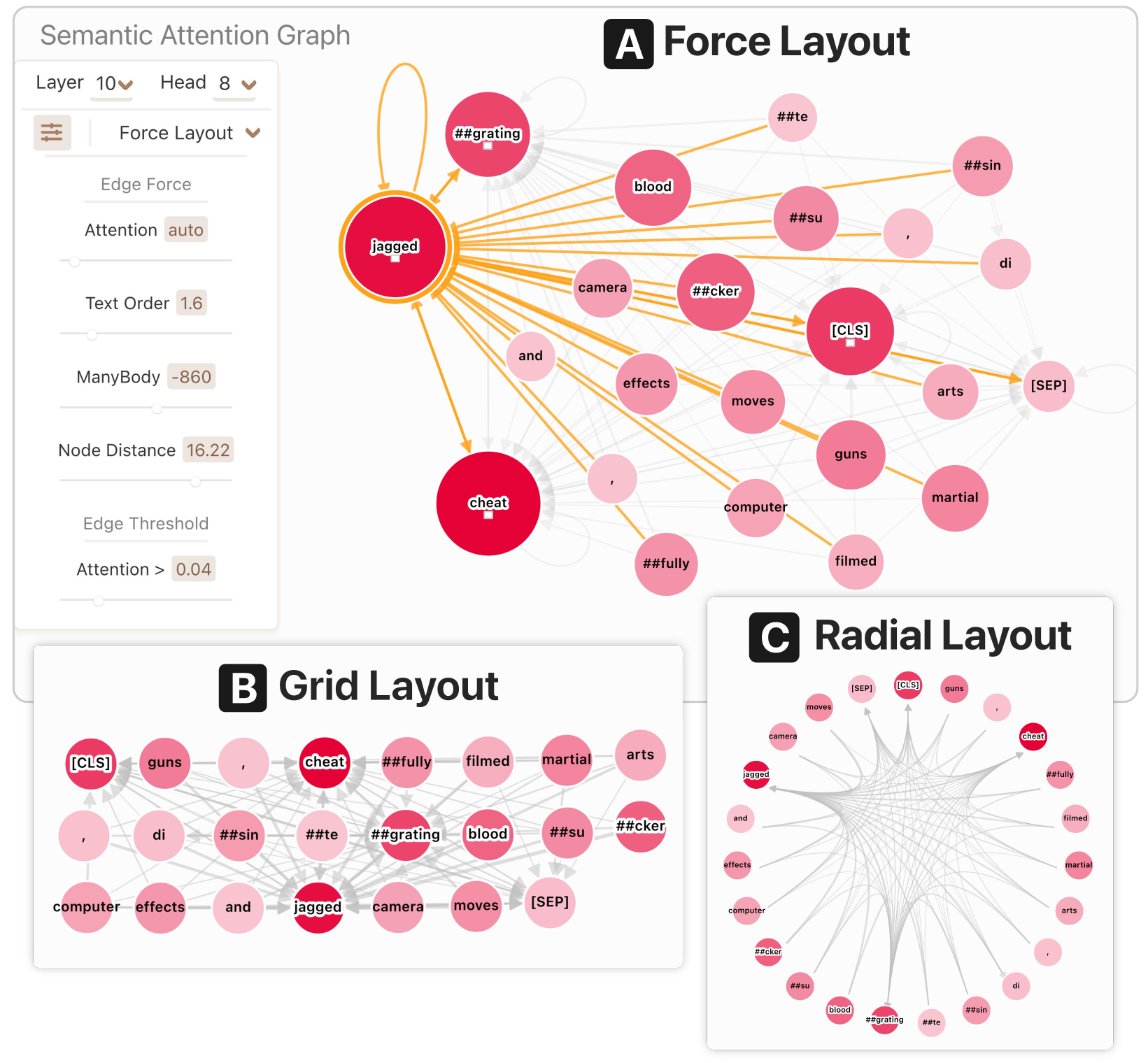}
    \caption{
    The \attentionMapView{} employs three graph visualization techniques to show the attention weights.
    (\textbf{A}) The \textit{force layout} allows users to flexibly change token positions;
    (\textbf{B}) the \textit{grid layout} enhances the readability of input sentence;
    (\textbf{C}) the \textit{radial layout} compactly highlights attention patterns. 
    }
    \label{fig:graph}
\end{figure}

\subsection{\attentionMapView{}}
The attention map at each head can be interpreted as an adjacency matrix, which can be visualized using different graph visualization techniques (\autoref{fig:graph}).
Users can primarily use this interactive graph view to inspect semantically significant attention heads, as defined the \overview{}.
Since the node color encodes the saliency score, linked to word's semantics~\cite{liVisualizingNeuralModels2016}, the behavior of the attention mechanism in the model can be evaluated from a semantic perspective.

Similarly to representations in the \expandedComparisonView{}, the \attentionMapView{} representations can be customized with interaction to allow for detailed attention inspection for selected tokens (\autoref{fig:graph}A), preserve token-order context in the \textit{Grid Layout} (\autoref{fig:graph}B), or allow for clear attention analysis in the \textit{Radial Layout} (\autoref{fig:graph}C).
Adjusting graph parameters in the side panel of this view encourages the user to customize the graph representation to ease attention analysis (eg. adjusting the \textit{edge threshold} parameter will only show attention weights with a greater magnitude) (\autoref{fig:graph}-A left).
We utilize linking to allow the user to interpret tokens in the context of their attention weights and dependence relations simultaneously as both nodes and edges are highlighted when a user hovers over a node in either the \attentionMapView{} or the \dependencyView{}.

\subsection{\instanceSelectionView{}}
For a robust understanding of the attention mechanisms in Transformers, it is important to explore the behavior of attention across interesting components of a sentence (eg. coreferences, word sense, etc.) present in various instances in a dataset.

The \textbf{\embeddingView{}} (\autoref{fig:appendix-instance}-A) uses UMAP~\cite{mcinnesUmap2018} to project text instance's model representation computed by concatenating the last four hidden state layers of BERT to a 2D space and visualizes it with a scatter plot.

The \textbf{\tableView{}} (\autoref{fig:appendix-instance}-B) allows for instance selection while providing the user with instance's true and predicted labels.
Users can hover over a dot in the \embeddingView{} to view the sentence text, and click a dot or a row in the \tableView{} to change \tool{}'s input sentence.
\section{Case Study}

\subsection{Understanding Sentiment in BERT}
How does a Transformer handle conflicting sentiment in opinionated phrases when resolving coreferences?
In \tool{}, we can explore the attention mechanism within a text instance from a movie review dataset, SST2~\cite{socherSST22013}, such as ``\textit{A coming-of-age film that avoids the cartoonish clichés and sneering humor of the genre as it provides a fresh view of an old type}."
Using this sentence, we can explore the concept of \textit{sentiment consistency} as proposed by~\cite{dingResolvingObjectCoreference2010} in the context of coreference resolution.

When interpreting the sentence above, it is clear to us that ``it" refers to the ``film" because the first half of the sentence expresses positive sentiment towards the ``film" and negative towards the ``genre," while the second half of the sentence represents a positive opinion on the ``film."
We can deduce that ``it" refers to the ``film" as sentiment is expressed in a consistent manner as discussed by~\cite{dingResolvingObjectCoreference2010}.
By exploring the \overview{} of \tool{} (\autoref{fig:appendix-sst}), we can select an attention head that conveys semantically significant information as indicated by the 2D color scale (eg. layer 1, head 7).
As we begin to analyze the \attentionMapView{} (\autoref{fig:appendix-sst}-left), we can hover over the node representing ``it" to visualize the attention behavior.
``It" attends highly to ``film," which validates the coreference resolution policy that we discussed above (\autoref{fig:appendix-sst}-right).
Users are encouraged to explore other attention heads as well to compare the behavior of the attention mechanism across various linguistic features.

\subsection{Penn Treebank Analysis}
Understanding attention across natural language tasks is pivotal for a systematic understanding of the attention mechanism as it relates to interpretability~\cite{vashishthAttentionInterpretabilityNLP2019}.
If we visualize BERT on a text corpus with annotated syntactic sentence structure, like Penn Treebank~\cite{marcusPennTreebank1993}, can attention accurately predict syntactic heads, and what patterns will we observe?

To investigate these ideas, we navigate to the \dependencyView{} within \tool{}.
Beginning in the \initialDependencyView{}, we observe edges of human annotated dependency relations connecting each token to its syntactic head, rather than part of speech (POS) tagging and dependency parsing annotations by the CoreNLP pipeline \cite{manningCoreNLP2014} when human annotations are not provided.
To identify whether some attention heads more accurately attend to the syntactic heads of each token, we will enter the \expandedComparisonView{} (\autoref{fig:comparison}) by clicking the \textit{Show Comparison} button in the toolbar.

As we see in \autoref{fig:comparison}-B3, \tool{} highlights correct syntactic head predictions by attention with a gradient edge, which is linked with the true dependencies in the \initialDependencyView{}.
After exploring various instances, we begin to understand patterns of certain attention heads.
For example, we observe that attention head 9 in layer 3 attends to nominals (group of nouns and adjectives: \texttt{obj}, \texttt{nmod}, \texttt{obl}, etc.) across unique instances (\autoref{fig:appendix-penn}).
This behavior highlights the syntax-aware attention that exists in BERT as discussed by~\cite{clarkWhatDoesBERT2019a}.
Visualizing consistent behavior by attention heads in Transformers outlines how the attention mechanism lends itself to model interpretability.

\subsection{Exploring DistilBERT}
The computational barrier to achieve state-of-the-art performance on natural language tasks with large pre-trained Transformers like BERT~\cite{delvinBert2019} was lowered when DistilBERT~\cite{sanhDistilBert2019}, a smaller version of BERT, was presented.
DistilBERT is 40\% smaller and retains up to 97\% performance compared to BERT with half as many self-attention layers.
With \tool{}, we can analyze attention mechanisms at various attention heads in DistilBERT to understand how attention compares to its larger version, BERT.

\begin{figure}[!tb]
    \centering
    \includegraphics[width=0.8\linewidth]{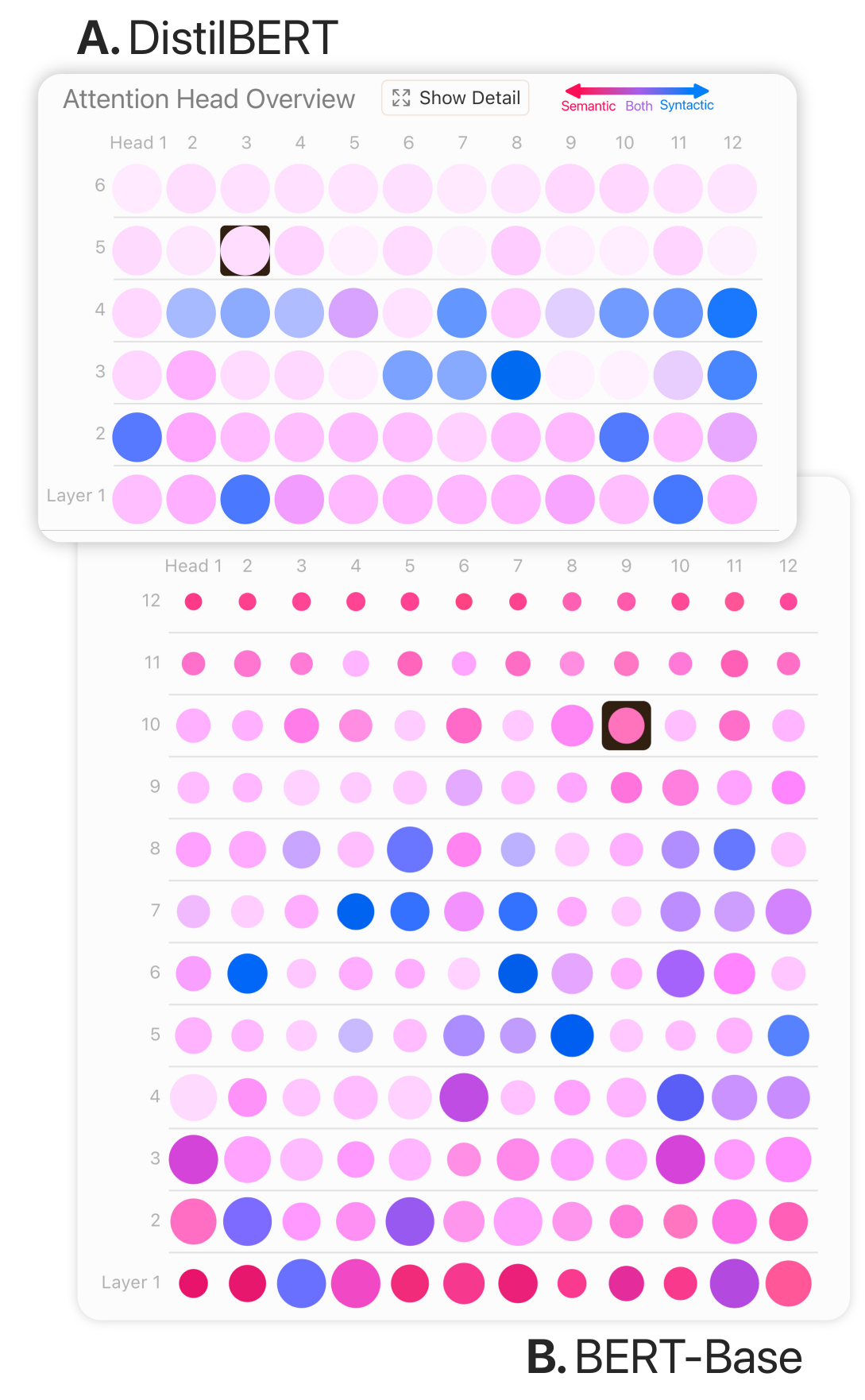}
    \caption{
    The \overview{} showing attention head roles for two transformer-based models.
    (\textbf{A}) All heads in DistilBERT are important and heads in early layers tend to have stronger linguistic alignment.
    (\textbf{B}) Attention heads in earlier layers tend to be more important and more semantic-aligned in BERT-Base.
    }
    \label{fig:distilbert}
\end{figure}

Using the \overview{} from \tool{} to visualize DistilBERT (\autoref{fig:distilbert}), we immediately notice that all radial attention head representations have the same diameter, unlike in the case of BERT.
Upon further inspection, we see that all attention heads have a confidence score that is very close to one via the tooltip present when hovering over an attention head, which indicates that every attention head has highly attended to tokens on average.
As we continue to explore the attention heads, we recognize a similar pattern of syntactic and semantic attention heads, but in the later layers the attention head rings have a much higher luminance in DistilBERT than they did in BERT.
According to the 2D color scale (\autoref{fig:overview}D), this represents a lower overall score meaning that these attention heads neither attend to primarily text semantics of grammatical structure.
It might imply that DistilBERT has learned some other linguistic knowledge beyond simple word semantics and syntactice dependencies.
We can then conduct quantitative experiment to test this hypothesis formed by using \tool{}.
\section{Discussion}

\tool{} aims to help NLP researchers and practitioners to explore attention mechanisms in transformer-based models with linguistic knowledge.
With overview + detail, linking + brushing, graph visualization techniques, \tool{} enables the users to investigate attention weights with different levels of abstraction in a context with both semantic and syntactic information.
Through use cases, we demonstrate that \tool{} not only helps users validate existing research results regarding the connections between attention weights with linguistic information, but also inspires the users to form hypothesis regarding the behavior and roles of attention heads across different models.

We acknowledge that there is an active discussion on whether attention weights can help people interpret transformer-based models \cite{jainAttentionNotExplanation2019b} and whether the attentions can be directly linked to the corresponding tokens in interpretation tasks \cite{brunnerIdentifiabilityTransformers2020}.
Our work joins the growing research body in NLP interpretability and human-centered NLP, highlighting novel visualization designs that can be generalized to other interactive NLP systems.
Despite the increasing popularity of applying Human-computer Interaction techniques to help people from various fields interact with complex NLP systems, little work have been done to evaluate how effective these tools are \cite{wangPuttingHumansNatural2021}.
To fill this research gap, we plan to run a user study to evaluate the usability and usefulness of \tool{}.
\section{Conclusion}

We present \tool{}, an interactive visualization system that fosters the exploration of the attention mechanism in transformer-based models with linguistic knowledge.
Through analysis from the model to the attention head level, users can explore how attention differs across a complex, state-of-the-art architecture over any instance within a dataset.
Our tool runs in modern web browsers and is open-sourced, broadening the public’s access to modern AI techniques.
We hope our work will inspire further research in understanding attention mechanisms and development of visualization tools that help people interact with complex NLP models.\looseness=-1

\section{Broader Impact}

We designed \tool{} with good intentions --- to help researchers and practitioners more easily explore attention weights in transformer-based models and investigate why their models succeed and fail.
However, bad actors could exploit this knowledge of whether and how the models may perform under different situations for malevolent purposes, such as manipulating the model prediction by injecting arbitrary keywords \cite{kurita-etal-2020-weight}.
The potential vulnerability warrants further study.

\section*{Acknowledgments}

We thank Haekyu Park, Rahul Duggal, and Nilaksh Das for their constructive feedback.
This work was supported in part by NSF grants IIS-1563816, CNS-1704701, DARPA GARD, gifts from Intel, NVIDIA, Bosch, Google, Amazon.
Use, duplication, or disclosure is subject to the restrictions as stated in Agreement number HR00112030001 between the Government and the Performer. 
\bibliographystyle{acl_natbib}
\bibliography{dodrio}

\clearpage

\section{Appendix}

\setcounter{figure}{0}
\renewcommand{\thetable}{S\arabic{table}}
\renewcommand{\thefigure}{S\arabic{figure}}

\noindent\begin{minipage}{\textwidth}
    \centering
    \includegraphics[width=\linewidth]{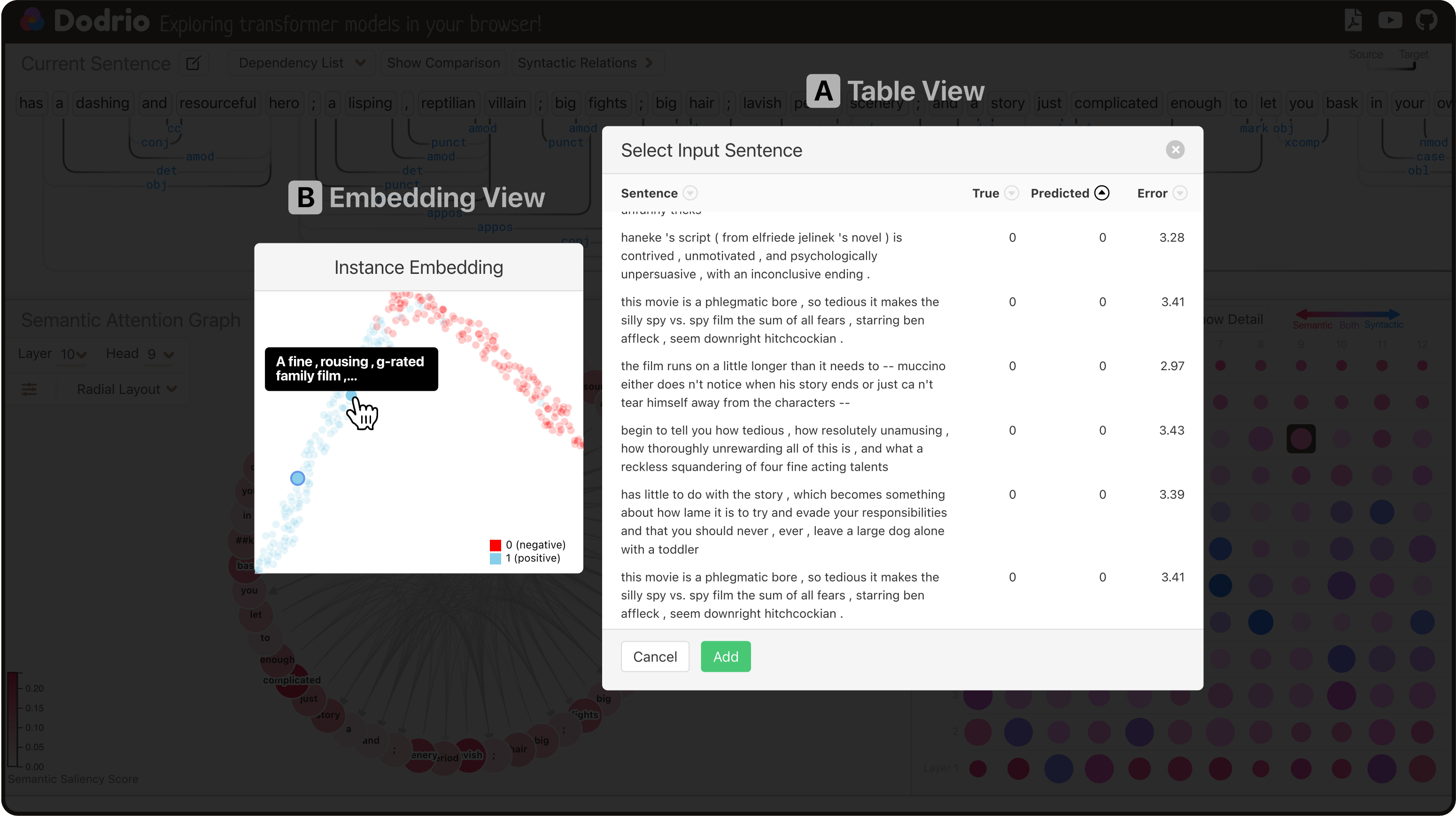}
    \captionof{figure}{
    The \instanceSelectionView{} within \tool{} encourages users to explore sentences with interesting linguistic features to understand how various attention heads throughout a model attend to them.
    (\textbf{A}) \textbf{Table View} presents all text instances in a tabular format with other dataset and task-specific information as well with sortable columns for efficient instance browsing.
    (\textbf{B}) \textbf{Embedding View} motivates users to inspect text clustered by dataset label to explore semantically interesting phrases.
    These views are linked, so that clicking an instance in either view will update the state of the other view, while setting the instance will update the global state of the entire interface. 
    }
    \label{fig:appendix-instance}
\end{minipage}

\vspace{8mm}

\noindent\begin{minipage}{\textwidth}
    \centering
    \includegraphics[width=\linewidth]{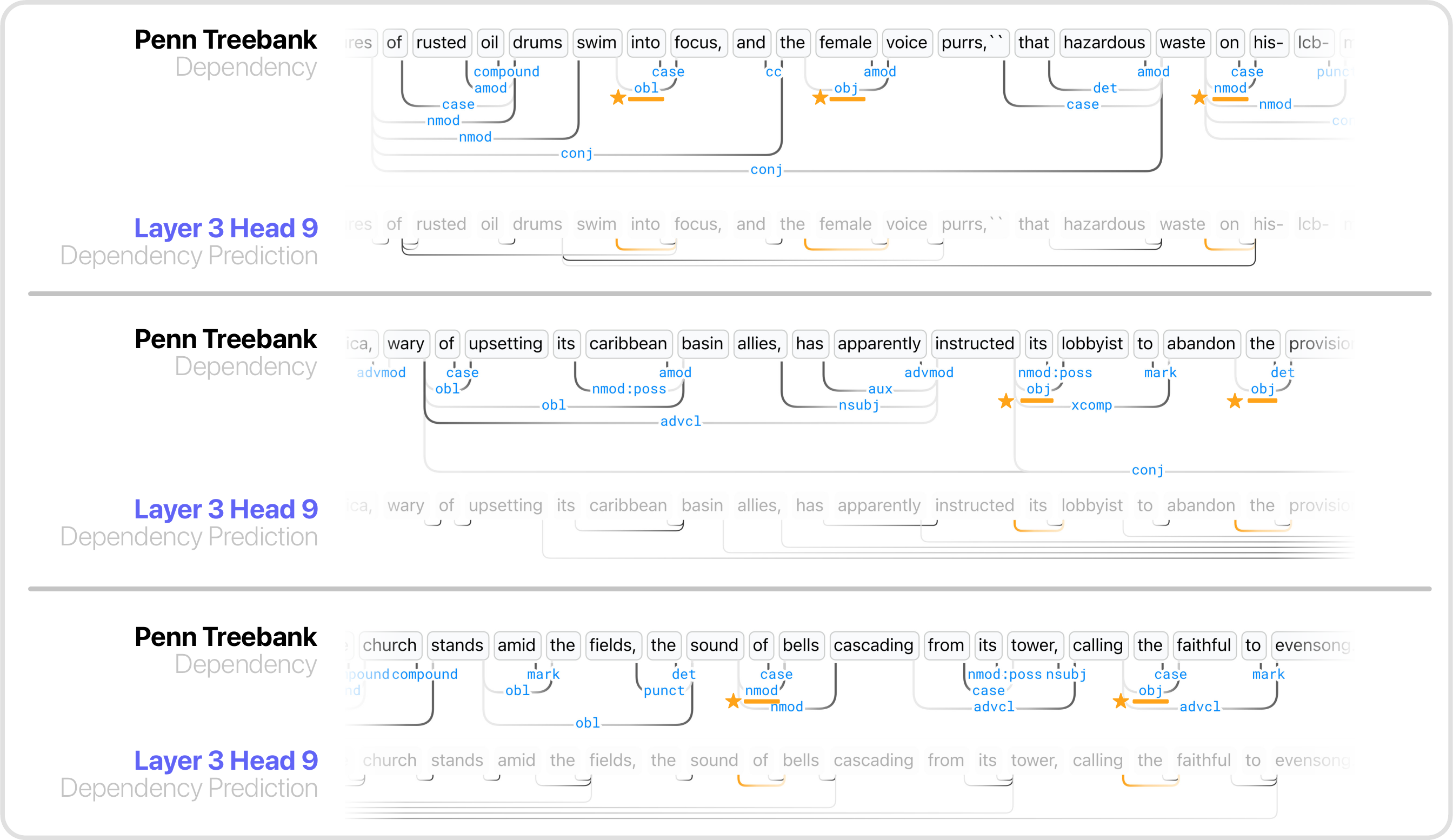}
    \captionof{figure}{
    The \expandedComparisonView{} visualizes syntactic relationships on the Penn Treeback dataset.
    It highlights attention head (Layer 3 Head 9) that can accurately predict the  nominal relationships (group of nouns and adjectives: \texttt{obj}, \texttt{nmod}, \texttt{obl}, etc.) across multiple unique instances.
    }
    \label{fig:appendix-penn}
\end{minipage}

\clearpage

\noindent\begin{minipage}{\textwidth}
    \centering
    \includegraphics[width=\linewidth]{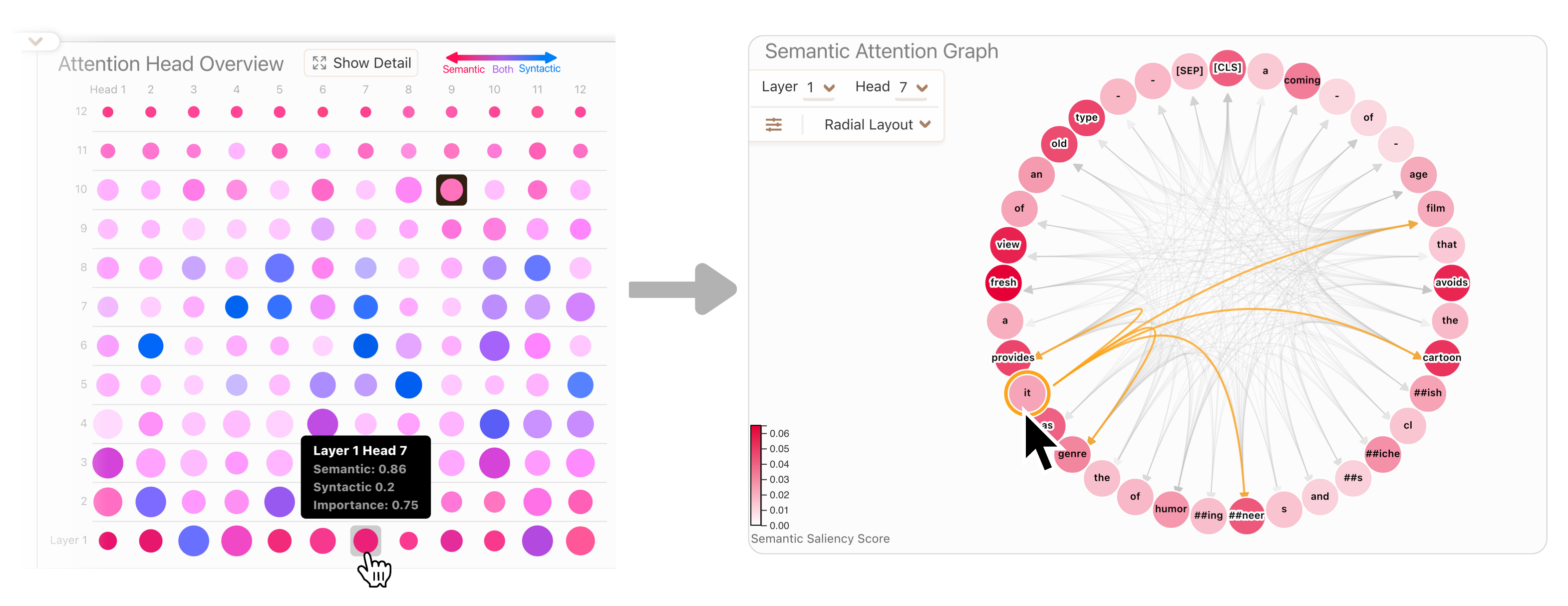}
    \captionof{figure}{
    The \textbf{\overview{}} (left) helps users identify interesting attention heads (e.g., more semantic-aligned and important heads), and then the \textbf{\attentionMapView{}} (right) quickly visualizes the attention weight pattern of the selected head on the current input sentence, allowing users to rapidly validate their hypothesis regarding attention head's linguistic knowledge.
    }
    \label{fig:appendix-sst}
\end{minipage}

\end{document}